\documentclass{article}

\usepackage[final]{neurips_2022_ml4ps}

\usepackage[utf8]{inputenc} 
\usepackage[T1]{fontenc}    
\usepackage{hyperref}       
\usepackage{url}            
\usepackage{booktabs}       
\usepackage{amsfonts}       
\usepackage{nicefrac}       
\usepackage{microtype}      
\usepackage{xcolor}         

\usepackage[pdftex]{graphicx}

\title{DS-GPS : A Deep Statistical Graph Poisson Solver \\ (for faster CFD simulations)}

\author{
    Matthieu Nastorg \\
    Université Paris-Saclay \\ 
    CNRS, Inria, LISN \\
    IFP Énergies nouvelles \\
    \texttt{matthieu.nastorg@inria.fr}
    \And
    Marc Schoenauer \\
    Université Paris-Saclay \\ 
    CNRS, Inria, LISN \\
    91405, Orsay, France \\
    \And 
    Guillaume Charpiat \\
    Université Paris-Saclay \\ 
    CNRS, Inria, LISN \\
    91405, Orsay, France \\
    \And 
    Thibault Faney \\
    IFP Énergies nouvelles \\
    92852, Rueil-Malmaison, France \\
    \And
    Jean-Marc Gratien \\
    IFP Énergies nouvelles \\
    92852, Rueil-Malmaison, France \\
    \And
    Michele-Alessandro Bucci \\
    Université Paris-Saclay \\ 
    CNRS, Inria, LISN \\
    91405, Orsay, France \\
}

\begin{document}

\maketitle

\begin{abstract}
This paper proposes a novel Machine Learning-based approach to solve a Poisson problem with mixed boundary conditions. Leveraging Graph Neural Networks, we develop a model able to process unstructured grids with the advantage of enforcing boundary conditions by design. By directly minimizing the residual of the Poisson equation, the model attempts to learn the physics of the problem without the need for exact solutions, in contrast to most previous data-driven processes where the distance with the available solutions is minimized.
\end{abstract}

\section{Introduction}

Numerical simulation of fluid flows plays an essential role in modelling many physical phenomena, such as climate, aerodynamics, electronics and medicine. Fluid dynamics can be described by the Navier–Stokes equations, but solving these equations at scale remains daunting, limited by the computational cost of resolving the smallest spatio-temporal scales. This is particularly true when dealing with incompressible Navier-Stokes equations. The well-known splitting method \citet{high-order-splitting-methods}, indeed, requires the costly resolution of a Poisson problem to compute the pressure field that guarantees the incompressibility constraint. Despite the significant progress made in the HPC community, the solution to the Poisson Pressure problem remains the major bottleneck in the speedup of CFD numerical simulations. Nowadays, data-driven methods, especially those based on deep neural networks, are reshaping the domain of numerical simulation. Neural networks can provide faster predictions, reducing turnaround time for workflows in engineering and science (see, for example, \citet{latent-space-physics} or \citet{cd-rom}). However, the lack of guarantees about the consistency and convergence of deep learning methods makes it impossible to implement deep learning models in the design and production stage of new engineering solutions. For this reason, recent studies propose to leverage the high flexibility of neural networks to alleviate specific and demanding computational operations in CFD simulations. 

A first attempt in this direction has been proposed in \citet{poisson-cnn}, where a Convolutional Neural Network (CNN) is trained to solve the inverse Poisson problem through supervised learning. In \citet{performance-and-accuracy-assessments}, a strategy based on first principles is adopted to train a CNN to minimize the error on the divergence of the velocity field to enforce the incompressibility constraint. Both approaches show promising results, but their application remains limited to structured grids with uniform discretization. Moreover, boundary conditions are learned by introducing an additional loss function or with an ad-hoc extra model, resulting in poor generalization performances for unseen configurations. To address these shortcomings, recent works have shown the ability of Graph Neural Networks (GNNs) to learn from unstructured meshes and accurately predict the dynamics of physical systems, as demonstrated in \cite{learning-mesh-based-simulation-with-graph-networks} and \citet{simulating-continuum-mechanics}. These approaches however are still based on supervised learning and require ground truth solutions.

Our work is related to \citet{deep-statistical-solver}, where the authors use GNNs to solve a Poisson problem with Dirichlet boundary conditions. We extend this work by proposing a new GNN-based architecture for solving iteratively a Poisson equation with mixed boundary conditions, thus making our model extensible to CFD cases. The physical state is first embedded in a latent space and then propagated through the mesh by combining message-passing mechanisms with a recurrent process, which significantly reduces the model size while allowing an "infinite" number of iterations until the problem converges. The method is trained in an end-to-end process minimizing the residual of the discretized Poisson problem in an entirely unsupervised manner. Consequently, it attempts to learn the physics of the equation and has the advantage of being trained without the need to compute ground truth values. By considering a node-specific architecture, the model respects the boundary conditions by design, avoiding the definition of an extra supervised loss that would make the model case-dependent. Although the approach to minimize the residual equation is well known since the PINNs \citet{pinns}, our method is distinguished by its ability to generalize to different domains, boundary conditions and initial solutions.

\section{Methodology}

\subsection{Problem statement}\label{prb-statement}
    
    Let $\Omega \subset \mathbb{R}^N$ be an open and bounded domain with smooth boundary $\partial \Omega$, $f$ be a continuous function defined on $\Omega$ and $g$ a continuous function defined on $\partial \Omega$. We consider the resolution of a Poisson problem with mixed boundary conditions (i.e. Dirichlet and homogeneous Neumann boundary conditions), which consists in finding a real-valued function $u$, defined on $\Omega$, solution of:
    
    \begin{equation}
        \left \{
        \begin{array}{rcl}
            -\Delta u &=& f \qquad \in \Omega \\
            u &=& g \qquad \in \partial \Omega_D \\
            \frac{\partial u }{\partial n} &=& 0 \qquad \in \partial \Omega_N
        \end{array}
        \right.
        \label{poisson-eq}
    \end{equation}
    
    where $n$ denotes the outward normal vector and $\partial \Omega = \partial \Omega_D \cup \partial \Omega_N$.
    
    To solve such a problem, the geometry of the domain is first discretized into an unstructured mesh, denoted $\Omega_h$. Equation \ref{poisson-eq} is then spatially discretized into a linear system
    
    \begin{equation}
        AU = B
        \label{linear-system}
    \end{equation} 
    
    using the Finite Element Method (FEM) with first order $P1$ finite elements. Thus, if $n$ denotes the number of degrees of freedom, it matches the number of nodes in $\Omega_h$. In \ref{linear-system}, $A \in \mathbb{R}^{n\times n}$ is the discretization of the Laplace operator, $B \in \mathbb{R}^n$ is the right-hand-side vector of size $n$ and $U \in \mathbb{R}^n$ is the solution vector to be sought.

    Let $\mathcal{F}$ be a set of continuous functions on $\Omega$ and $\mathcal{G}$ a set of continuous functions on $\partial \Omega$. We denote as $\mathcal{P}$ a set of Poisson problems, parametrized by $p \in \Sigma$, such that any element $E_p \in \mathcal{P}$ is described as a triplet $E_p = \left( \Omega_{p}, ~f_p, ~g_p \right)$ where $\Omega_p \subset \Omega$, $f_p \in \mathcal{F}$ and $g_p \in \mathcal{G}$. For all $p \in \Sigma$, we consider $E_{h,p} \in \mathcal{P}_h$, the discretization of $E_p$, such that $E_{h,p} = \left(\Omega_{h,p}, ~A_p, ~B_p \right)$ where $\Omega_{h,p} \subset \Omega_h$ and $A_p$ and $B_p$ are extracted from the FEM linear system \ref{linear-system}. 

    The idea of this work is to build a Machine Learning solver, parametrized by a vector $\theta \in \Theta$, which outputs the 
    solution $U_p$ of a discretized Poisson problem $E_{h,p} \in \mathcal{P}_h$ : 
    
    \begin{equation}
        U_p ~ = ~ \textit{solver}_\theta\left(E_{h,p}\right) ~ = ~ \textit{solver}_\theta\left(\Omega_{h,p}, ~A_p, ~B_p\right) 
    \end{equation}
    
    \subsection{Graph formulation and statistical problem}
    
    In \ref{linear-system}, it should be noted that the structure of matrix $A$ encodes the geometry of its corresponding mesh. Indeed, for each node, using first-order finite elements leads to the creation of local stencils, which represent local connections between mesh nodes.
    
    Therefore, a discretized Poisson problem $E_h = \left(\Omega_h, ~A, ~B \right)$ can then be interpreted as a graph problem $G = (n, ~A, ~B)$ where $n$ is the number of nodes in the graph, $A = (a_{ij})_{i,j \in[n]} \in \mathbb{R}$ is the weighted adjacency matrix that represents the interactions between the nodes and $B = (b_i)_{i\in[n]} \in \mathbb{R}$ is defined as some local external inputs. Vector $U = (u_i)_{i\in[n]} \in \mathbb{R}$ represents the state of the graph where $u_i$ is the state of the node $i$. We denote by $\mathcal{S}$ the set of all such graphs $G$ and $\mathcal{U}$ the set of all such states $U$. In our case, we have $\mathcal{S} = \mathcal{P}_h$. 
    
    Additionally, we define $\mathcal{L}$, a real-valued function defined on a pair $(U,G)$ such that :
    
    \begin{equation}
         \mathcal{L}(U,G) = ||AU - B||^2 = \displaystyle \sum_{i \in [n]}(-b_i + \sum_{j \in [n]} a_{i,j}u_j)^2
         \label{loss-function}
    \end{equation}
    
    The basic problem is given a graph $G$, to find an optimal state in $\mathcal{U}$ that minimizes \ref{loss-function}. More generally, we are interested in building a solver, parametrized by $\theta \in \Theta$, which finds such an optimal state for many graphs $G$, sampled from a given distribution $\mathcal{D}$ over $\mathcal{S}$. Hence, the statistical problem can be formulated as : 
    
    Given a distribution $\mathcal{D}$ on space $\mathcal{S}$ and a loss function $\mathcal{L}$, solve :
    
    \begin{equation}
        \theta^* = \textit{argmin}_{\theta \in \Theta} \mathbb{E}_{G \sim \mathcal{D}} \left[\mathcal{L}\left(\textit{solver}_{\theta}\left(G\right),G\right)\right]
        \label{ssp-problem}
    \end{equation}

    A crucial upside using GNNs is related to the treatment of boundary conditions. In \ref{linear-system}, the Dirichlet boundary conditions break the symmetry of the matrix $A$. Therefore, we use an undirected graph at those particular nodes, sending information only to its neighbours without receiving any. Conversely, Neumann nodes have bi-directional edges with specific message passing based on satisfying local Neumann boundary equations.    
        
\subsection{Architecture description}

    The proposed architecture is a recurrent process that acts on a d-dimensional latent space $\mathcal{H}$ during $k$ fixed iterations.
        
    At the beginning of the process, a multilayer perceptron (MLP) maps the initial solution to an initial latent state. This trainable function can be interpreted as an encoder projecting the physical space $\mathcal{U}$ to a higher dimensional latent space $\mathcal{H}$ on which we apply GNN layers.
        
    Then, a recurrent process is considered for computing the next latent state while considering a specific treatment for each type of node: i) for an \textit{Interior} node, two message passing are computed using geometric information (e.g., distances between nodes) to account for edge directionality. These messages concatenated with the discretized value of the force function $f$ are then passed as input to a GRU cell whose output is the next desired latent state ii) for a \textit{Dirichlet} boundary node, the value is preserved across iterations iii) for a \textit{Neumann} boundary node, a specific message passing is computed to output the next latent state that respects, by design, this specific boundary condition.

    At each iteration, an MLP decodes the intermediate latent state back into the physical space. For each intermediate state, two losses are computed. The first one is the quadratic norm of the residual equation (i.e. equation \ref{loss-function}). The second one is the loss corresponding to the encoding-decoding process. The final training loss is calculated as a cumulative sum of all intermediate losses.
        
    It is important to emphasize that constructing matrices $A$ and $B$ in \ref{linear-system} is only necessary for training the model. Indeed, during inference, the model only requires as input the graph (and its relative geometric information) and the discretized value of the force function. We always ensure that the initially provided solution already satisfies the Dirichlet boundary conditions.  

\section{Results}

To build the dataset, we consider 2-dimensional random geometries whose coordinates are in the unit square. Each geometry is discretized into an unstructured triangular mesh with approximately $300$ to $600$ nodes. Functions $f$ and $g$ are defined as polynomials with coefficients sampled from a uniform distribution. Therefore the dataset has different sources of randomness that define $\Sigma$ in \ref{prb-statement}: i) shape of the mesh ii) volumetric forcing iii) Dirichlet boundary conditions. The number $k$ of iterations for training the model is fixed to $20$, matching the average diameter of the considered meshes to propagate the information across all the nodes correctly.

\begin{figure}[!ht]
\centering
\includegraphics[width=1.0\linewidth]{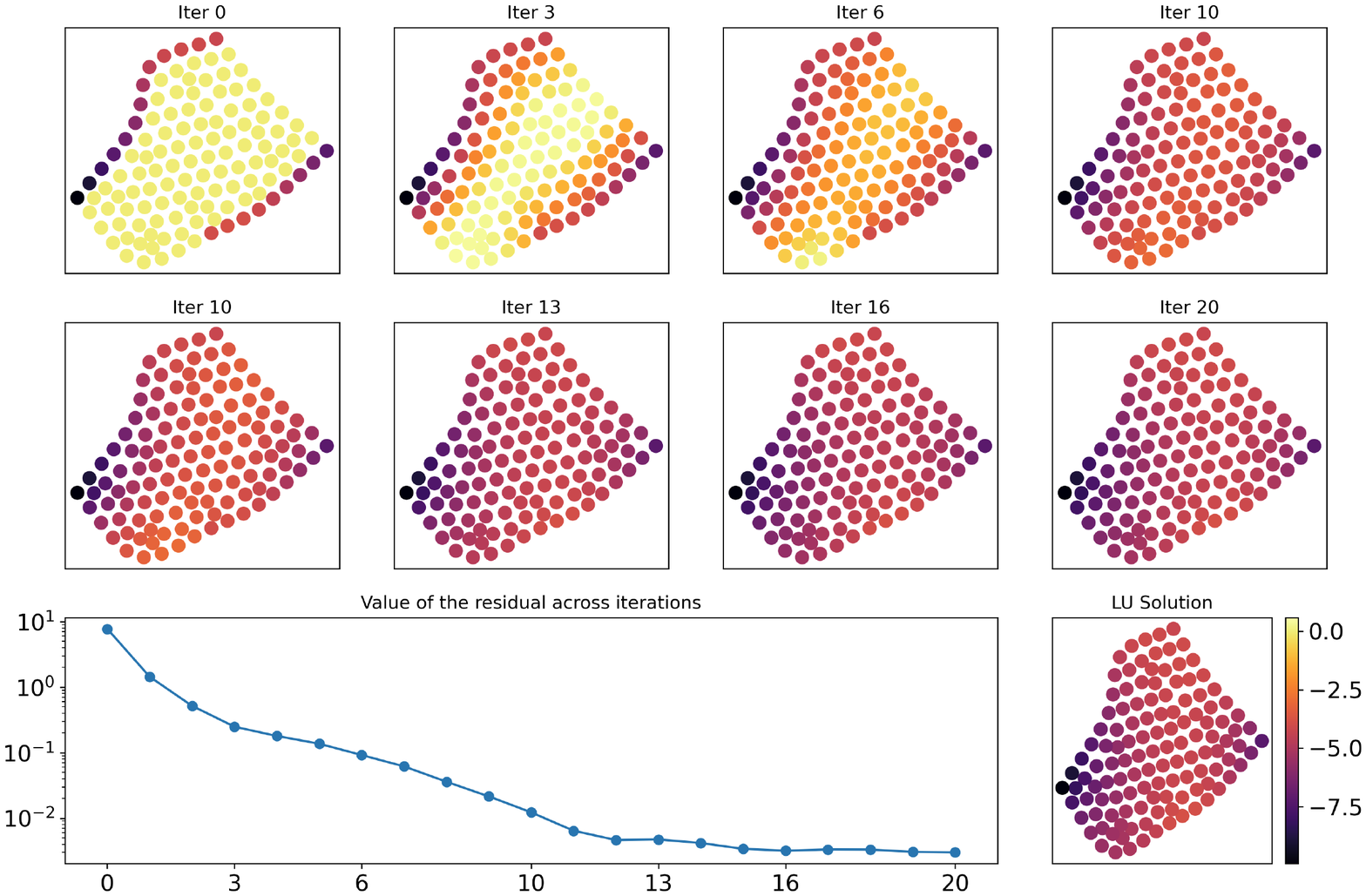}
\caption{Resolution of a Poisson pressure problem with the proposed data-driven method}
\label{fig:my_label}
\end{figure}

Figure \ref{fig:my_label} displays the evolution of a discretized Poisson problem solution extracted from the test set using the proposed data-driven. The solution is initialized to zero except on the Dirichlet boundary (top left plot) to be compared to the numerical solution obtained with the LU decomposition of the discretized operator (bottom right plot). As expected, the information is propagated from the Dirichlet boundary nodes until the convergence of the problem. At the bottom left of figure \ref{fig:my_label}, the $l_2$-norm of the residual loss across different iterations is reported, achieving a value of order $10^{-3}$ at the final stage. Additionally, the $l_2$-norm of the difference between the true solution and the predicted one (i.e. the solution from the last iteration) is computed, leading to an error of magnitude $10^{-2}$. By taking advantage of GPU parallelization, we observe that our method can compute the solution ten times faster than LU decomposition. Finally, it is essential to point out that the current model has about $4500$ parameters, a significant decrease from the method of \citet{deep-statistical-solver}, in which there are about $50000$ weights. 

\section{Conclusion}

This paper proposes a novel Machine Learning-based approach to solve a Poisson problem with mixed boundary conditions. Leveraging GNNs and a node-specific architecture, our method extends to any mesh size and satisfies boundary conditions by design. By directly minimizing the residual of the discretized Poisson problem, our model learns the physics of the equation (i.e. the discretized operator) and can output a solution ten times faster than traditional solvers.

Numerical experiments show that the model converges to a fixed point for a sufficient number of fixed iterations. Future work will then focus on combining this GNN approach with the theory of implicit layers (see \citet{deq} and \citet{deq-mutliscale}). The improved model would offer convergence guarantees while automatically determining the correct number of iterations to perform, thus correcting two major downsides of the proposed method.

Future assessment of the model performance will be done by direct integration in a CFD solver to boost the fractional step scheme. Two configurations will be considered. Firstly, the GNN model will be used as a preconditioner for standard iterative methods (e.g. Jacobi, Gauss-Seidel). Secondly, an exclusive deep learning solution will be considered to mimic multigrid methods.

\medskip

\section{Potential broader impact}

This paper presents a new Machine Learning-based algorithm to solve a Poisson problem with mixed boundary conditions. This work is, so far, developed in a very general and theoretical framework without focusing on a specific application. Therefore, we believe that the potential for abuse of our work is minimal. However, solving and accelerating computational fluid dynamics problems could have many positive or negative impacts. Thus, this work and any related topics should be used cautiously to avoid abuse in the Machine Learning or Numerical Simulation community.

\bibliography{sample}

\section*{Checklist}




\begin{enumerate}

\item For all authors...
\begin{enumerate}
    \item Do the main claims made in the abstract and introduction accurately reflect the paper's contributions and scope? \answerYes{}
    \item Did you describe the limitations of your work? \answerYes{}
    \item Did you discuss any potential negative societal impacts of your work? \answerYes{}
    \item Have you read the ethics review guidelines and ensured that your paper conforms to them? \answerYes{}
\end{enumerate}

\item If you are including theoretical results...
\begin{enumerate}
    \item Did you state the full set of assumptions of all theoretical results? \answerNA{}
    \item Did you include complete proofs of all theoretical results? \answerNA{}
\end{enumerate}

\item If you ran experiments...
\begin{enumerate}
    \item Did you include the code, data, and instructions needed to reproduce the main experimental results (either in the supplemental material or as a URL)? \answerNA{}
    \item Did you specify all the training details (e.g., data splits, hyperparameters, how they were chosen)? \answerNA{}
    \item Did you report error bars (e.g., with respect to the random seed after running experiments multiple times)? \answerNA{}
    \item Did you include the total amount of compute and the type of resources used (e.g., type of GPUs, internal cluster, or cloud provider)? \answerNA{}
\end{enumerate}

\item If you are using existing assets (e.g., code, data, models) or curating/releasing new assets...
\begin{enumerate}
      \item If your work uses existing assets, did you cite the creators? \answerNA{}
      \item Did you mention the license of the assets? \answerNA{}
      \item Did you include any new assets either in the supplemental material or as a URL? \answerNA{}
      \item Did you discuss whether and how consent was obtained from people whose data you're using/curating? \answerNA{}
      \item Did you discuss whether the data you are using/curating contains personally identifiable information or offensive content? \answerNA{}
\end{enumerate}

\item If you used crowdsourcing or conducted research with human subjects...
\begin{enumerate}
      \item Did you include the full text of instructions given to participants and screenshots, if applicable? \answerNA{}
      \item Did you describe any potential participant risks, with links to Institutional Review Board (IRB) approvals, if applicable? \answerNA{}
      \item Did you include the estimated hourly wage paid to participants and the total amount spent on participant compensation? \answerNA{}
\end{enumerate}

\end{enumerate}

\end{document}